\documentclass[10pt,twocolumn,letterpaper]{article}

\usepackage[pagenumbers]{cvpr} 

\usepackage[dvipsnames]{xcolor}

\usepackage{graphicx}
\usepackage{amsmath}
\usepackage{amssymb}
\usepackage{booktabs}
\usepackage[ruled,vlined,noend]{algorithm2e}

\definecolor{cvprblue}{rgb}{0.21,0.49,0.74}
\usepackage[pagebackref,breaklinks,colorlinks,citecolor=cvprblue]{hyperref}

\def\paperID{13027} 
\def\confName{CVPR}
\def\confYear{2024}

\title{HiFi Tuner: \\
High-Fidelity Subject-Driven Fine-Tuning for Diffusion Models}

\author{Zhonghao Wang$^{1,2}$, Wei Wei$^{4}$, Yang Zhao$^{1}$, Zhisheng Xiao$^{1}$,\\
Mark Hasegawa-Johnson$^{2}$, Humphrey Shi$^{2,3}$, Tingbo Hou$^{1}$\\
{\small $^1$Google, $^2$UIUC, $^3$Georgia Tech, $^4$Accenture}}

\begin{document}
\maketitle
\begin{abstract}
This paper explores advancements in high-fidelity personalized image generation through the utilization of pre-trained text-to-image diffusion models. While previous approaches have made significant strides in generating versatile scenes based on text descriptions and a few input images, challenges persist in maintaining the subject fidelity within the generated images. In this work, we introduce an innovative algorithm named HiFi Tuner to enhance the appearance preservation of objects during personalized image generation. Our proposed method employs a parameter-efficient fine-tuning framework, comprising a denoising process and a pivotal inversion process. Key enhancements include the utilization of mask guidance, a novel parameter regularization technique, and the incorporation of step-wise subject representations to elevate the sample fidelity. Additionally, we propose a reference-guided generation approach that leverages the pivotal inversion of a reference image to mitigate unwanted subject variations and artifacts. We further extend our method to a novel image editing task: substituting the subject in an image through textual manipulations. Experimental evaluations conducted on the DreamBooth dataset using the Stable Diffusion model showcase promising results. Fine-tuning solely on textual embeddings improves CLIP-T score by 3.6 points and improves DINO score by 9.6 points over Textual Inversion. When fine-tuning all parameters, HiFi Tuner improves CLIP-T score by 1.2 points and improves DINO score by 1.2 points over DreamBooth, establishing a new state of the art. 
\end{abstract}

\section{Introduction}
\label{sec:intro}

\begin{figure}[t]
\centering
\includegraphics[width=0.47\textwidth]{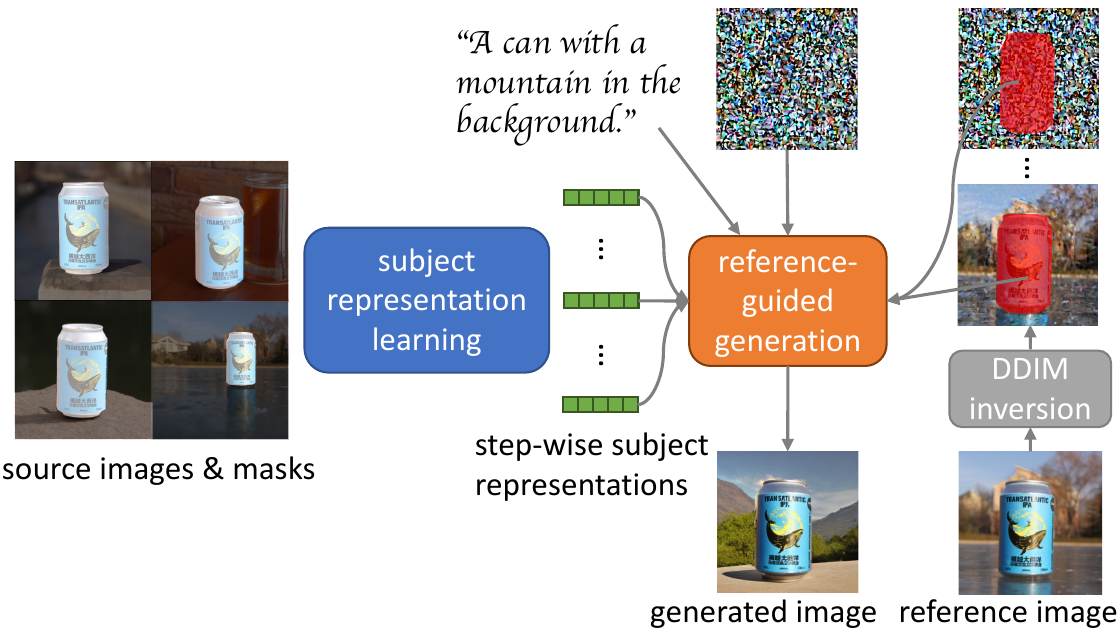}
\caption{\small{Illustration of HiFi Tuner. We first learn the step-wise subject representations with subject source images and masks. Then we select and transform the reference image, and use DDIM inversion to obtain its noise latent trajectory. Finally, we generate an image controlled by the prompt, the step-wise subject representations and the reference subject guidance.}}
\label{pics:teaser}
\end{figure}

Diffusion models \cite{sohl2015deep,ho2020denoising} have demonstrated a remarkable success in producing realistic and diverse images. The advent of large-scale text-to-image diffusion models \cite{saharia2022photorealistic, rombach2022high, ramesh2022hierarchical}, leveraging expansive web-scale training datasets \cite{schuhmann2022laion, chen2022pali}, has enabled the generation of high-quality images that align closely with textual guidance. Despite this achievement, the training data remains inherently limited in its coverage of all possible subjects. Consequently, it becomes infeasible for diffusion models to accurately generate images of specific, unseen subjects based solely on textual descriptions. As a result, personalized generation has emerged as a pivotal research problem. This approach seeks to fine-tune the model with minimal additional costs, aiming to generate images of user-specified subjects that seamlessly align with the provided text descriptions.

We identify three drawbacks of existing popular methods for subject-driven fine-tuning \cite{ruiz2023dreambooth, hu2022lora, gal2023an, ruiz2023hyperdreambooth}. Firstly, a notable imbalance exists between sample quality and parameter efficiency in the fine-tuning process. For example, Textual Inversion optimizes only a few parameters in the text embedding space, resulting in poor sample fidelity. Conversely, DreamBooth achieves commendable sample fidelity but at the cost of optimizing a substantial number of parameters. Ideally, there should be a parameter-efficient method that facilitates the generation of images with satisfactory sample fidelity while remaining lightweight for improved portability. Secondly, achieving a equilibrium between sample fidelity and the flexibility to render objects in diverse scenes poses a significant challenge. Typically, as fine-tuning iterations increase, the sample fidelity improves, but the flexibility of the scene coverage diminishes. Thirdly, current methods struggle to accurately preserve the appearance of the input object. Due to the extraction of subject representations from limited data, these representations offer weak constraints to the diffusion model. Consequently, unwanted variations and artifacts may appear in the generated subject. 

In this study, we introduce a novel framework named HiFi Tuner for subject fine-tuning that prioritizes the parameter efficiency, thereby enhancing sample fidelity, preserving the scene coverage, and mitigating undesired subject variations and artifacts. Our denoising process incorporates a mask guidance to reduce the influence of the image background on subject representations. Additionally, we introduce a novel parameter regularization method to sustain the model's scene coverage capability and design a step-wise subject representation mechanism that adapts to parameter functions at different denoising steps. We further  propose a reference-guided generation method that leverages pivotal inversion of a reference image. By integrating guiding information into the step-wise denoising process, we effectively address issues related to unwanted variations and artifacts in the generated subjects. Notably, our framework demonstrates versatility by extending its application to a novel image editing task: substituting the subject in an image with a user-specified subject through textual manipulations. 

We summarize the contributions of our work as follows. Firstly, we identify and leverage three effective techniques to enhance the subject representation capability of textual embeddings. This improvement significantly aids the diffusion model in generating samples with heightened fidelity. Secondly, we introduce a novel reference-guided generation process that successfully addresses unwanted subject variations and artifacts in the generated images. Thirdly, we extend the application of our methodology to a new subject-driven image editing task, showcasing its versatility and applicability in diverse scenarios. Finally, we demonstrate the generic nature of HiFi Tuner by showcasing its effectiveness in enhancing the performance of both the Textual Inversion and the DreamBooth. 

\section{Related Works}
\label{sec:rel_works}
\textbf{Subject-driven text-to-image generation}. This task requires the generative models generate the subject provided by users in accordance with the textual prompt description. Pioneer works \cite{casanova2021instance, nitzan2022mystyle} utilize Generative Adversarial Networks (GAN) \cite{goodfellow2020generative} to synthesize images of a particular instance. Later works benefit from the success of diffusion models \cite{rombach2022high, saharia2022photorealistic} to achieve a superior faithfulness in the personalized generation. Some works \cite{chen2023reimagen, sheynin2022knn} rely on retrieval-augmented architecture to generate rare subjects. However, they use weakly-supervised data which results in an unsatisfying faithfullness for the generated images. There are encoder-based methods \cite{chen2023subject, jia2023taming, shi2023instantbooth} that encode the reference subjects as a guidance for the diffusion process. However, these methods consume a huge amount of time and resources to train the encoder and does not perform well for out-of-domain subjects. Other works \cite{ruiz2023dreambooth, gal2023an} fine-tune the components of diffusion models with the provided subject images. Our method follows this line of works as our models are faithful and generic in generating rare and unseen subjects.

\textbf{Text-guided image editing}. This task requires the model to edit an input image according to the modifications described by the text. Early works \cite{patashnik2021styleclip, gal2023an} based on diffusion models \cite{rombach2022high, saharia2022photorealistic} prove the effectiveness of manipulating textual inputs for editing an image. Further works \cite{avrahami2022blended, meng2022sdedit} propose to blend noise with the input image for the generation process to maintain the layout of the input image. Prompt-to-Prompt \cite{hertz2023prompttoprompt, mokady2023null} manipulates the cross attention maps from the image latent to the textual embedding to edit an image and maintain its layout. InstructPix2Pix \cite{brooks2023instructpix2pix} distills the diffusion model with image editing pairs synthesized by Prompt-to-Prompt to implement the image editing based on instructions.

\section{Methods}
\label{sec:methods}

\begin{figure*}[t]
\centering
\includegraphics[width=0.99\textwidth]{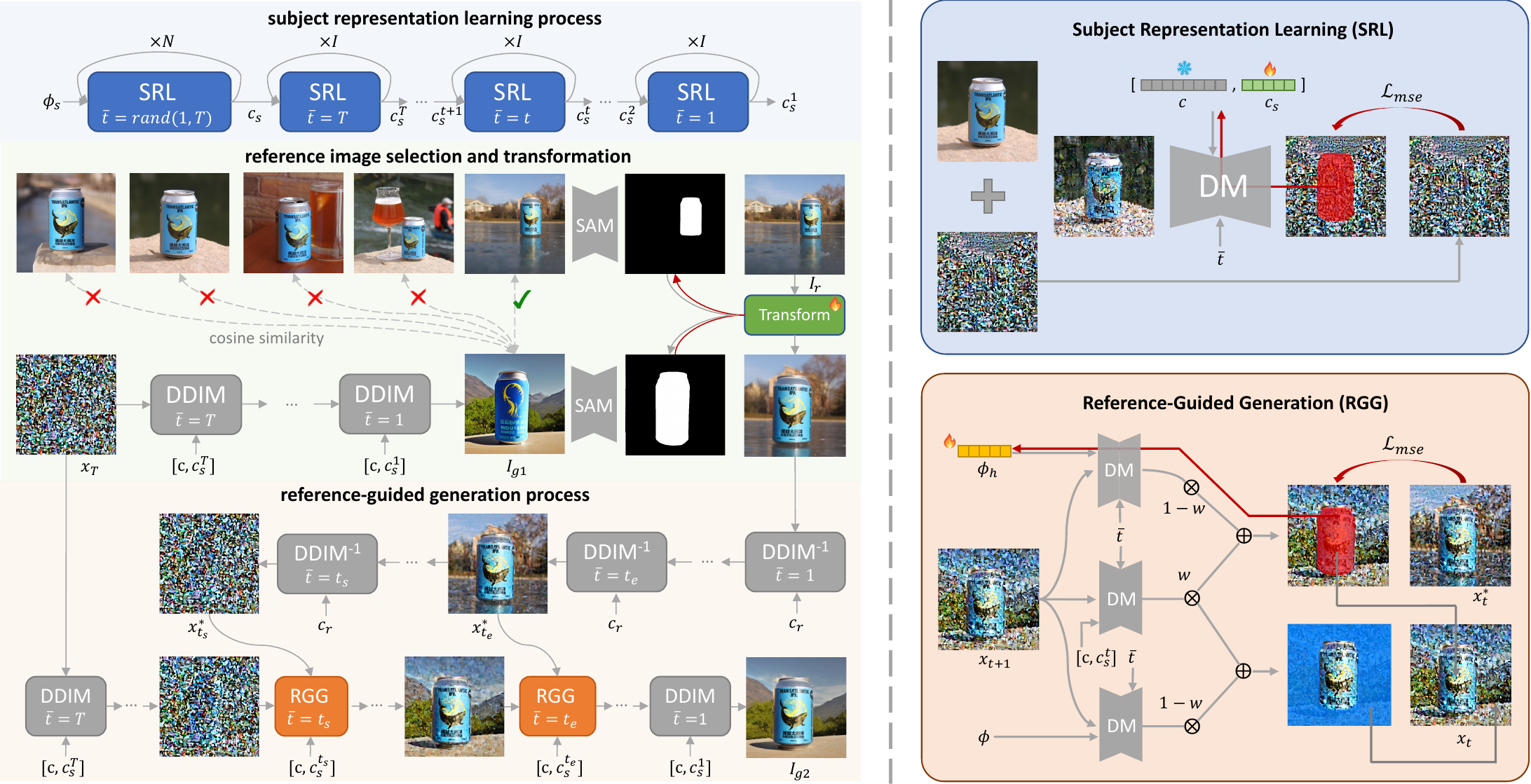}
\caption{\small{The framework of HiFi Tuner. The grey arrows stand for the data flow direction. The red arrows stand for the gradient back propagation direction. $SAM$ stands for the Segment Anything \cite{kirillov2023segany} model. $DM$ stands for the Stable Diffusion \cite{rombach2022high} model. $DDIM$ and ${DDIM}^{-1}$ stands for the DDIM denoising step and inversion step respectively.}}
\vspace{-3mm}
\label{pics:framework}
\end{figure*}

In this section, we elaborate HiFi Tuner in details. We use the denoising process to generate subjects with appearance variations and the inversion process to preserve the details of subjects. In section \ref{subsec:backgrounds}, we present some necessary backgrounds for our work. In section \ref{subsec:denoise}, we introduce the three proposed techniques that help preserving the subject identity. In section \ref{subsec:guided_infer}, we introduce the reference-guided generation technique, which merits the image inversion process to further preserve subject details. In section \ref{subsec:subject_rep}, we introduce an extension of our work on a novel image editing application -- personalized subject replacement with only textual prompt edition. 

\subsection{Backgrounds}
\label{subsec:backgrounds}

\textbf{Stable diffusion} \cite{rombach2022high} is a widely adopted framework in the realm of text-to-image diffusion models. Unlike other methods \cite{saharia2022photorealistic, ramesh2022hierarchical}, Stable diffusion is a latent diffusion model, where the diffusion model is trained within the latent space of a Variational Autoencoder (VAE).  To accomplish text-to-image generation, a text prompt undergoes encoding into textual embeddings $c$ using a CLIP text encoder\cite{radford2021learning}. Subsequently, a random Gaussian noise latent $x_T$ is initialized. The process then recursively denoises noisy latent $x_t$ through a noise predictor network $\epsilon_\theta$ with the conditioning of $c$. Finally, the VAE decoder is employed to project the denoised latent $x_0$ onto an image. During the sampling process, a commonly applied mechanism involves classifier-free guidance \cite{ho2022classifier} to enhance sample quality. Additionally, deterministic samplers, such as DDIM \cite{song2020denoising}, are employed to improve sampling efficiency.  The denoising process can be expressed as
\begin{equation}
\begin{split}
    x_{t-1} &= F^{(t)}(x_t, c, \phi) \\
            &= \beta_t x_t - \gamma_t (w\epsilon_\theta(x_t, c) + (1-w)\epsilon_\theta(x_t, \phi)).
    \label{eqn:ddim_denoise}
\end{split}
\end{equation}
where $\beta_t$ and $\gamma_t$ are time-dependent constants; $w$ is the classifier-free guidance weight; $\phi$ is the CLIP embedding for a null string.

\textbf{Textual inversion} \cite{gal2023an}. As a pioneer work in personalized generation, Textual Inversion introduced the novel concept that a singular learnable textual token is adequate to represent a subject for the personalization. Specifically, the method keeps all the parameters of the diffusion model frozen, exclusively training a word embedding vector $c_s$ using the diffusion objective:
\begin{align}
    \mathfrak{L}_s(c_s) = \min_{c_s} \|\epsilon_\theta (x_t, [c, c_s]) - \epsilon \|_2^2,
    \label{eqn:textual_inv_loss}
\end{align}
where $[c, c_s]$ represents replacing the object-related word embedding in the embedding sequence of the training caption (\eg ``a photo of A") with the learnable embedding $c_s$. After $c_s$ is optimized, this work applies $F^{(t)}(x_t, [c, c_s], \phi)$ for generating personalized images from prompts. 

\textbf{Null-text inversion} \cite{mokady2023null} method introduces an inversion-based approach to image editing, entailing the initial inversion of an image input to the latent space, followed by denoising with a user-provided prompt. This method comprises two crucial processes: a pivotal inversion process and a null-text optimization process. The pivotal inversion involves the reversal of the latent representation of an input image, denoted as $x_0$, back to a noise latent representation, $x_T$, achieved through the application of reverse DDIM. This process can be formulated as reparameterizing Eqn. (\ref{eqn:ddim_denoise}) with $w=1$:
\begin{equation}
    x_{t+1} = {F^{-1}}^{(t)}(x_t, c) = \overline{\beta_t} x_t + \overline{\gamma_t} \epsilon_\theta(x_t, c)
    \label{eqn:pivotal_inversion}
\end{equation}
We denote the latent trajectory attained from the pivotal inversion as $[x_0^*,...,x_T^*]$. However, naively applying Eqn. (\ref{eqn:ddim_denoise}) for $x_T^*$ will not restore $x_0^*$, because $\epsilon_\theta(x_t, c) \neq \epsilon_\theta(x_{t-1}^*, c)$. To recover the original image, Null-text inversion trains a null-text embedding $\phi_t$ for each timestep $t$ force the the denoising trajectory to stay close to the forward trajectory $[x_0^*,...,x_T^*]$. The learning objective is
\begin{align}
    \mathfrak{L}_h^{(t)}(\phi_t) = \min_{\phi_t} \|x_{t-1}^* - F^{(t)}(x_t, c, \phi_t) \|_2^2.
    \label{eqn:null_optim}
\end{align}
After training, image editing techniques such as the prompt-to-prompt \cite{hertz2023prompttoprompt} can be applied with the learned null-text embeddings $\{\phi_t^*\}$ to allow manipulations of the input image. 
\subsection{Learning subject representations}
\label{subsec:denoise}
We introduce three techniques for improved learning of the representations that better capture the given object.

\textbf{Mask guidance} One evident issue we observed in Textual Inversion is the susceptibility of the learned textual embedding, $c_s$, to significant influence from the backgrounds of training images. This influence often imposes constraints on the style and scene of generated samples and makes identity preservation more challenging due to the limited capacity of the textual embedding, which is spent on unwanted background details. We present a failure analysis of Textual Inversion in the Appendix ~\ref{app:TI_failure_analysis}. 
To address this issue, we propose a solution involving the use of subject masks to confine the loss during the learning process of $c_s$. This approach ensures that the training of $c_s$ predominantly focuses on subject regions within the source images. Specifically, binary masks of the subjects in the source images are obtained using Segment Anything (SAM) \cite{kirillov2023segany}, an off-the-shelf instance segmentation model. The Eqn.~(\ref{eqn:textual_inv_loss}) is updated to a masked loss: 
\begin{equation}
    \mathfrak{L}_s(c_s) = \min_{c_s} \| M \odot (\epsilon_\theta (x_t, [c, c_s]) - \epsilon) \|_2^2,
    \label{eqn:mask_loss}
\end{equation}
where $\odot$ stands for element-wise product, and $M$ stands for a binary mask of the subject. This simple technique mitigates the adverse impact of background influences and enhancing the specificity of the learned textual embeddings.

\textbf{Parameter regularization} We aim for the learned embedding, $c_s$, to obtain equilibrium between identity preservation and the ability to generate diverse scenes. To achieve this balance, we suggest initializing $c_s$ with a portion of the null-text embedding, $\phi_s$, and introducing an L2 regularization term. This regularization term is designed to incentivize the optimized $c_s$ to closely align with $\phi_s$:
\begin{equation}
\small
    \mathfrak{L}_s(c_s) = \min_{c_s} \| M \odot (\epsilon_\theta (x_t, [c, c_s]) - \epsilon) \|_2^2 + w_s \|c_s - \phi_s\|_2^2.
    \label{eqn:reg_loss}
\end{equation}
Here, $c_s \in \mathbb{R}^{n \times d} $ where $n$ is the number of tokens and $d$ is the embedding dimension, and $w_s$ is a regularization hyper-parameter. We define $\phi_s$ as the last $n$ embeddings of $\phi$ and substitute the last $n$ embeddings in $c$ with $c_s$, forming $[c, c_s]$. It is noteworthy that $[c, c_s] = c$ if $c_s$ is not optimized, given that $\phi$ constitutes the padding part of the embedding. This regularization serves two primary purposes. Firstly, the stable diffusion model is trained with a $10\%$ caption drop, simplifying the conditioning to $\phi$ and facilitating classifier-free guidance \cite{ho2022classifier}. Consequently, $\phi$ is adept at guiding the diffusion model to generate a diverse array of scenes, making it an ideal anchor point for the learned embedding. Secondly, due to the limited data used for training the embedding, unconstrained parameters may lead to overfitting with erratic scales. This overfitting poses a risk of generating severely out-of-distribution textual embeddings.

\begin{figure}[t]
\centering
\includegraphics[width=0.47\textwidth]{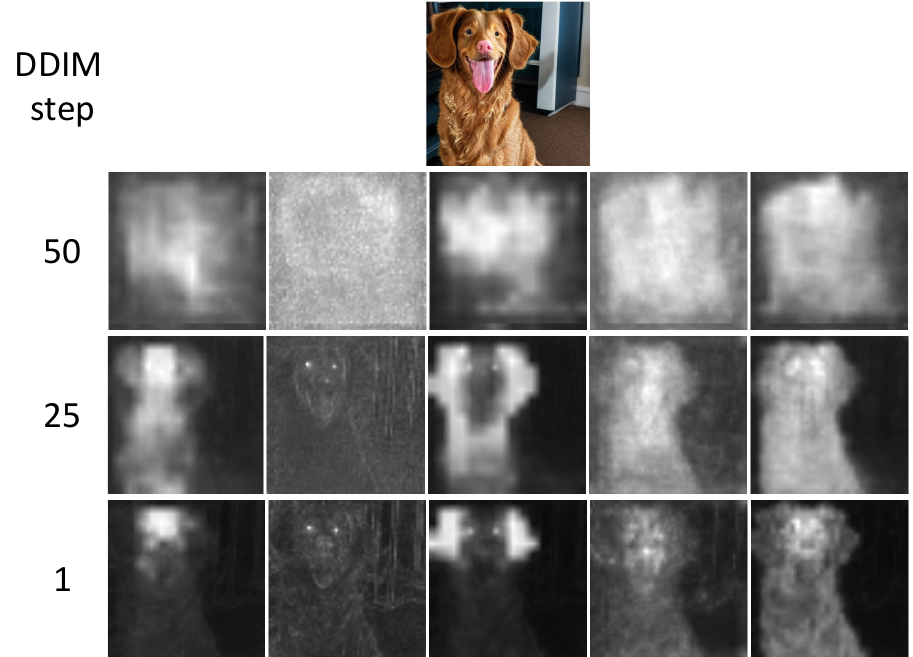}
\caption{\small{Step-wise function analysis of $c_s$. We generate an image from a noise latent with DDIM and an optimized $c_s$ representing a subject dog. The text prompt is "A sitting dog". The top image is the result generated image. We follow \cite{hertz2023prompttoprompt} to obtain the attention maps with respect to the 5 token embeddings of $c_s$ as shown in the below images. The numbers to the left refer to the corresponding DDIM denoising steps. In time step 50, the 5 token embeddings of $c_s$ are attended homogeneously across the latent vectors. In time step 1, these token embeddings are attended mostly by the subject detailed regions such as the forehead, the eyes, the ears, \etc}}
\label{pics:step_attns}
\end{figure}

\textbf{Step-wise subject representations} We observe that the learned textual embedding, $c_s$, plays distinct roles across various denoising time steps. It is widely acknowledged that during the sampling process. In early time steps where $t$ is large, the primary focus is on generating high-level image structures, while at smaller values of $t$, the denoising process shifts its emphasis toward refining finer details. Analogous functional distinctions exist for the role of $c_s$. Our analysis of $c_s$ across time steps, presented in Fig. \ref{pics:step_attns}, underscores these variations.
Motivated by this observation, we propose introducing time-dependent embeddings, $c_s^t$, at each time step instead of a single $c_s$ to represent the subject. This leads to a set of embeddings, $[c_s^1,...,c_s^T]$, working collectively to generate images. To ensure smooth transitions between time-dependent embeddings, we initially train a single $c_s$ across all time steps. Subsequently, we recursively optimize ${c_s^t}$ following DDIM time steps, as illustrated in Algorithm \ref{algr:stepwise_cs}. This approach ensures that $c_s^t$ is proximate to $c_s^{t+1}$ by initializing it with $c_s^{t+1}$ and optimizing it for a few steps. After training, we apply 
\begin{equation}
    x_{t-1} = F^{(t)}(x_t, [c, c_s^t], \phi)
    \label{eqn:stepwise_denoise}
\end{equation}
with the optimized $[c_s^1,...,c_s^T]$ to generate images.  


\begin{algorithm}[t]
\SetAlgoLined
\small
 \caption{\small{Optimization algorithm for $c_s^t$. $T$ is DDIM time steps. $I$ is the optimization steps per DDIM time step. $X_0$ is the set of encoded latents of the source images. $N_s(\cdot)$ is the DDIM noise scheduler. $\mathfrak{L}_s(\cdot)$ refers to the loss function in Eqn. (\ref{eqn:reg_loss}).}}
 \label{algr:stepwise_cs}
\KwResult{$C_s$}
 $C_s=\{\}$, $c_s^{T+1}=c_s$\\
 \For{$t = [T,...,1]$}{
    $c_s^t = c_s^{t+1}$ \\
    \For{$i = [1,...,I]$}{
        $\epsilon \sim \mathcal{N}(0, 1)$, $x_0 \in X_0$, $x_t = N_s(x_0, \epsilon, t)$\\
        $c_s^t = c_s^t - \eta \nabla_{c_s^t} \mathfrak{L}_s(c_s^t)$
    }
    $C_s= C_s \cup \{c_s^t\}$
 }
\end{algorithm}

\subsection{Reference-guided generation}
\label{subsec:guided_infer}
Shown in Figure \ref{pics:framework}, we perform our reference-guided generation in three steps. First, we determine the initial latent $x_T$ and follow the DDIM denoising process to generate an image. Thus, we can determine the subject regions of $\{x_t\}$ requiring guiding information and the corresponding reference image. Second, we transform the reference image and inverse the latent of the transformed image to obtain a reference latent trajectory, $[x_0^*,...,x_T^*]$. Third, we start a new denoising process from $x_T$ and apply the guiding information from $[x_0^*,...,x_T^*]$ to the guided regions of $\{x_t\}$. Thereby, we get a reference-guided generated image.

\textbf{Guided regions and reference image}. First, we determine the subject regions of $x_t$ that need the guiding information. Notice that $x_t \in \mathbb{R}^{H\times W \times C}$, where $H$, $W$ and $C$ are the height, width and channels of the latent $x_t$ respectively. Following the instance segmentation methods \cite{he2017mask, liu2021swin}, we aim to find a subject binary mask $M_g$ to determine the subset $x_t^s \in \mathbb{R}^{m \times C}$ corresponding to the subject regions. Because DDIM \cite{song2020denoising} is a deterministic denoising process as shown in Eqn. (\ref{eqn:ddim_denoise}), once $x_T$, $c$ and $\phi$ are determined, the image to be generated is already determined. Therefore, we random initialize $x_T$ with Gaussian noise; then, we follow Eqn. (\ref{eqn:stepwise_denoise}) and apply the decoder of the stable diffusion model to obtain a generated image, $I_{g1}$; by applying Grounding SAM \cite{liu2023grounding, kirillov2023segany} with the subject name to $I_{g1}$ and resizing the result to $H \times W$, we obtain the subject binary mask $M_g$. Second, we determine the reference image by choosing the source image with the closest subject appearance to the subject in $I_{g1}$, since the reference-guided generation should modify $\{x_t\}$ as small as possible to preserve the image structure. As pointed out by DreamBooth~\cite{ruiz2023dreambooth}, DINO \cite{caron2021emerging} score is a better metric than CLIP-I \cite{radford2021learning} score in measuring the subject similarity between two images. Hence, we use ViT-S/16 DINO model \cite{caron2021emerging} to extract the embedding of $I_{g1}$ and all source images. We choose the source image whose DINO embedding have the highest cosine similarity to the DINO embedding of $I_{g1}$ as the reference image, $I_r$. We use Grounding SAM \cite{liu2023grounding, kirillov2023segany} to obtain the subject binary mask $M_r$ of $I_r$. 

\textbf{Reference image transformation and inversion}. First, we discuss the transformation of $I_r$. Because the subject in $I_{g1}$ and the subject in $I_r$ are spatially correlated with each other, we need to transform $I_r$ to let the subject better align with the subject in $I_{g1}$. As the generated subject is prone to have large appearance variations, it is noneffective to use image registration algorithms, e.g. RANSAC \cite{fischler1981random}, based on local feature alignment. We propose to optimize a transformation matrix
\begin{equation}
\footnotesize
    T_\theta = \begin{bmatrix}
        \theta_1 & 0 & 0 \\
        0 & \theta_1 & 0 \\
        0 & 0 & 1
    \end{bmatrix}
    \begin{bmatrix}
        \cos(\theta_2) & -\sin{\theta_2} & 0 \\
        \sin{\theta_2} & \cos(\theta_2) & 0 \\
        0 & 0 & 1
    \end{bmatrix}
    \begin{bmatrix}
        1 & 0 & \theta_3 \\
        0 & 1 & \theta_4 \\
        0 & 0 & 1
    \end{bmatrix}
\label{eqn:t_matrix}
\end{equation}
composed of scaling, rotation and translation such that $T_\theta(M_r)$ best aligns with $M_g$. Here, $\{\theta_i\}$ are learnable parameters, and $T_\theta(\cdot)$ is the function of applying the transformation to an image. $T_\theta$ can be optimized with
\begin{equation}
    \mathfrak{L}_t = \min_\theta \|T_\theta(M_r) - M_g\|_1^1.
    \label{eqn:t_matrix_loss}
\end{equation}
Please refer to the Appendix ~\ref{app:opt_T_theta} for a specific algorithm optimizing $T_\theta$. We denote the optimized $T_\theta$ as $T_\theta^*$ and the result of $T_\theta^*(M_r)$ as $M_r^*$. Thereafter, we can transform $I_r$ with $T_\theta^*(I_r)$ to align the subject with the subject in $I_{g1}$. Notice that the subject in $T_\theta^*(I_r)$ usually does not perfectly align with the subject in $I_{g1}$. A rough spatial location for placing the reference subject should suffice for the reference guiding purpose in our case. Second, we discuss the inversion of $T_\theta^*(I_r)$. We use BLIP-2 model \cite{li2023blip2} to caption $I_r$ and use a CLIP text encoder to encode the caption to $c_r$. Then, we encode $T_\theta^*(I_r)$ into $x_0^*$ with a Stable Diffusion image encoder. Finally, we recursively apply Eqn. (\ref{eqn:pivotal_inversion}) to obtain the reference latent trajectory, $[x_0^*,...,x_T^*]$.

\textbf{Generation process}. There are two problems with the reference-guided generation: 1) the image structure needs to be preserved; 2) the subject generated needs to conform with the context of the image. We reuse $x_T$ in step 1 as the initial latent. If we follow Eqn. (\ref{eqn:stepwise_denoise}) for the denoising process, we will obtain $I_{g1}$. We aim to add guiding information to the denoising process and obtain a new image $I_{g2}$ such that the subject in $I_{g2}$ has better fidelity and the image structure is similar to $I_{g1}$. Please refer to Algorithm \ref{algr:generate_process} for the specific reference-guided generation process. As discussed in Section \ref{subsec:denoise}, the stable diffusion model focuses on the image structure formation at early denoising steps and the detail polishing at later steps. If we incur the guiding information in early steps, $I_{g2}$ is subject to have structural change such that $M_r^*$ cannot accurately indicate the subject regions. It is harmful to enforce the guiding information at later steps either, because the denoising at this stage gathers useful information mostly from the current latent. Therefore, we start and end the guiding process at middle time steps $t_s$ and $t_e$ respectively. At time step $t_s$, we substitute the latent variables corresponding to the subject region in $x_t$ with those in $x_t^*$. We do this for three reasons: 1) the substitution enables the denoising process to assimilate the subject to be generated to the reference subject; 2) the latent variables at time step $t_s$ are close to the noise space so that they are largely influenced by the textual guidance as well;  3) the substitution does not drastically change the image structure because latent variables have small global effect at middle denoising steps. We modify Eqn. (\ref{eqn:null_optim}) to Eqn. (\ref{eqn:h_optim}) for guiding the subject generation.
\begin{equation}
\begin{split}
    \mathfrak{L}_h^{(t)}(\phi_h) = \min_{\phi_h} \|x_{t-1}^*[M_r^*] - 
     F^{(t)}(x_{t}, [c, c_s^t], \phi_h)[M_r^*] \|_2^2
    \label{eqn:h_optim}
\end{split}
\end{equation}
Here, $x_t[M]$ refers to latent variables in $x_t$ indicated by the mask $M$. Because $\phi_h$ is optimized with a few steps per denoising time step, the latent variables corresponding to the subject regions change mildly within the denoising time step. Therefore, at the next denoising time step, the stable diffusion model can adapt the latent variables corresponding to non-subject regions to conform with the change of the latent variables corresponding to the subject regions. Furthermore, we can adjust the optimization steps for $\phi_h$ to determine the weight of the reference guidance. More reference guidance will lead to a higher resemblance to the reference subject while less reference guidance will result in more variations for the generated subject.
\begin{algorithm}[t]
\SetAlgoLined
\small
 \caption{\small{Reference-guided generation algorithm. $J$ is the number of optimization steps for $\phi_h$ per denoising step. $\mathfrak{L}_h^{(t)}(\cdot)$ refers to the loss function in Eqn. (\ref{eqn:h_optim}).}}
 \label{algr:generate_process}
\KwResult{$x_0$}
\textbf{Inputs}: $t_s$, $t_e$, $x_T$, $M_r^*$, $c$, $\phi$, $[c_s^1,...,c_s^T]$, $[x_0^*,...,x_T^*]$  \\
\For{$t = [T,...,1]$}{
 \If{$t == t_s$}{
  $\phi_h = \phi$ \\
  $x_{t}[M_r^*] = x_{t}^*[M_r^*]$
 }
 $x_{t-1} = F^{(t)}(x_{t}, [c, c_s^t], \phi)$ \\
 \If{$t \leqslant t_s$ \textbf{and} $t \geqslant t_e$}{
  \For{$j = [1,...,J]$}{
   $\phi_h = \phi_h - \eta \nabla_{\phi_h} \mathfrak{L}_h^{(t)}(\phi_h)$
  }
  $x_{t-1}[M_r^*] = F^{(t)}(x_{t}, [c, c_s^t], \phi_h)[M_r^*]$
 }
}
\end{algorithm}
\subsection{Personalized subject replacement}
\label{subsec:subject_rep}
We aim to use the learned subject textual representations to replace the subject in an image with the user-specified subject. Although there are methods \cite{lugmayr2022repaint, xie2023smartbrush, zhang2023paste, li2023dreamedit} inpainting the image area with a user-specified subject, our method has two advantages over them. First, we do not specify the inpainting area of the image; instead, our method utilize the correlation between the textual embeddings and the latent variables to identify the subject area. Second, our method can generate a subject with various pose and appearance such that the added subject better conforms to the image context.

We first follow the fine-tuning method in Section \ref{subsec:denoise} to obtain the step-wise subject representations $[c_s^1,...,c_s^T]$. We encode the original image $I_r$ to $x_0^r$ with the Stable Diffusion image encoder; then we use BLIP-2 model \cite{li2023blip2} to caption $I_r$ and encode the caption into $c^r$ with the Stable Diffusion language encoder. We identify the original subject word embedding in $c^r$ and substitute that with the new subject word embedding $w_g$ to attain a $c^g$ (e.g. `cat' $\rightarrow$ `dog' in the sentence `a photo of siting cat'). Then we follow Algorithm \ref{algr:subject_replace} to generate the image with the subject replaced. Referring to the prompt-to-prompt paper \cite{hertz2023prompttoprompt}, we store the step-wise cross attention weights with regard to the word embeddings in $c^r$ to ${a_t^r}^*$. $A^{(t)}(\cdot, \cdot, \cdot)$ performs the same operations as $F^{(t)}(\cdot, \cdot, \cdot)$ in Eqn. (\ref{eqn:ddim_denoise}) but returns $x_{t-1}$ and ${a_t^r}^*$. We also modify $F^{(t)}(\cdot, \cdot, \cdot)$ to $\Tilde{F}_{[c_s^t, w_g]}^{(t)}(\cdot, \cdot, \cdot, {a_t^r}^*)$ such that all token embeddings use fixed cross attention weights ${a_t^r}^*$ except that $[c_s^t, w_g]$ use the cross attention weights of the new denoising process.
\begin{algorithm}[t]
\SetAlgoLined
\small
 \caption{\small{Personalized subject replacement algorithm. ${F^{-1}}^{(t)}$ refers to Eqn. (\ref{eqn:pivotal_inversion}). $K$ is the optimization steps for null-text optimization. $\mathfrak{L}_h^{(t)}(\cdot)$ refers to Eqn. (\ref{eqn:null_optim})}}
 \label{algr:subject_replace}
\KwResult{$x_0^g$}
\textbf{Inputs}: $x_0^r$, $c^r$, $c^g$, $[c_s^1,...,c_s^T]$  \\
${x_0^r}^* = x_0^r$ \\
\For{$t = [0,...,T-1]$}{
 ${x_{t+1}^r}^* = {F^{-1}}^{(t)}({x_t^r}^*, c^r)$
}
$x_T^r={x_T^r}^*$, $\phi_T=\phi$ \\
\For{$t = [T,...,1]$}{
 \For{$k = [1,...,K]$}{
  $\phi_t = \phi_t - \eta \nabla_{\phi_t} \mathfrak{L}_h^{(t)}(\phi_t)$
 }
 $x_{t-1}^r, {a_t^r}^* = A^{(t)}(x_{t}^r, c^r, \phi_t)$ \\
 $\phi_{t-1} = \phi_t^* = \phi_t$ \\
}
$x_T^g={x_T^r}^*$ \\
\For{$t = [T,...,1]$}{
  $x_{t-1}^g = \Tilde{F}_{[c_s^t, w_g]}^{(t)}(x_t^g, [c^g, c_s^t], \phi_t^*, {a_t^r}^*)$
}
\end{algorithm}

\section{Experiments}
\label{sec:experiments}

\begin{figure*}[t]
\centering
\includegraphics[width=0.96\textwidth]{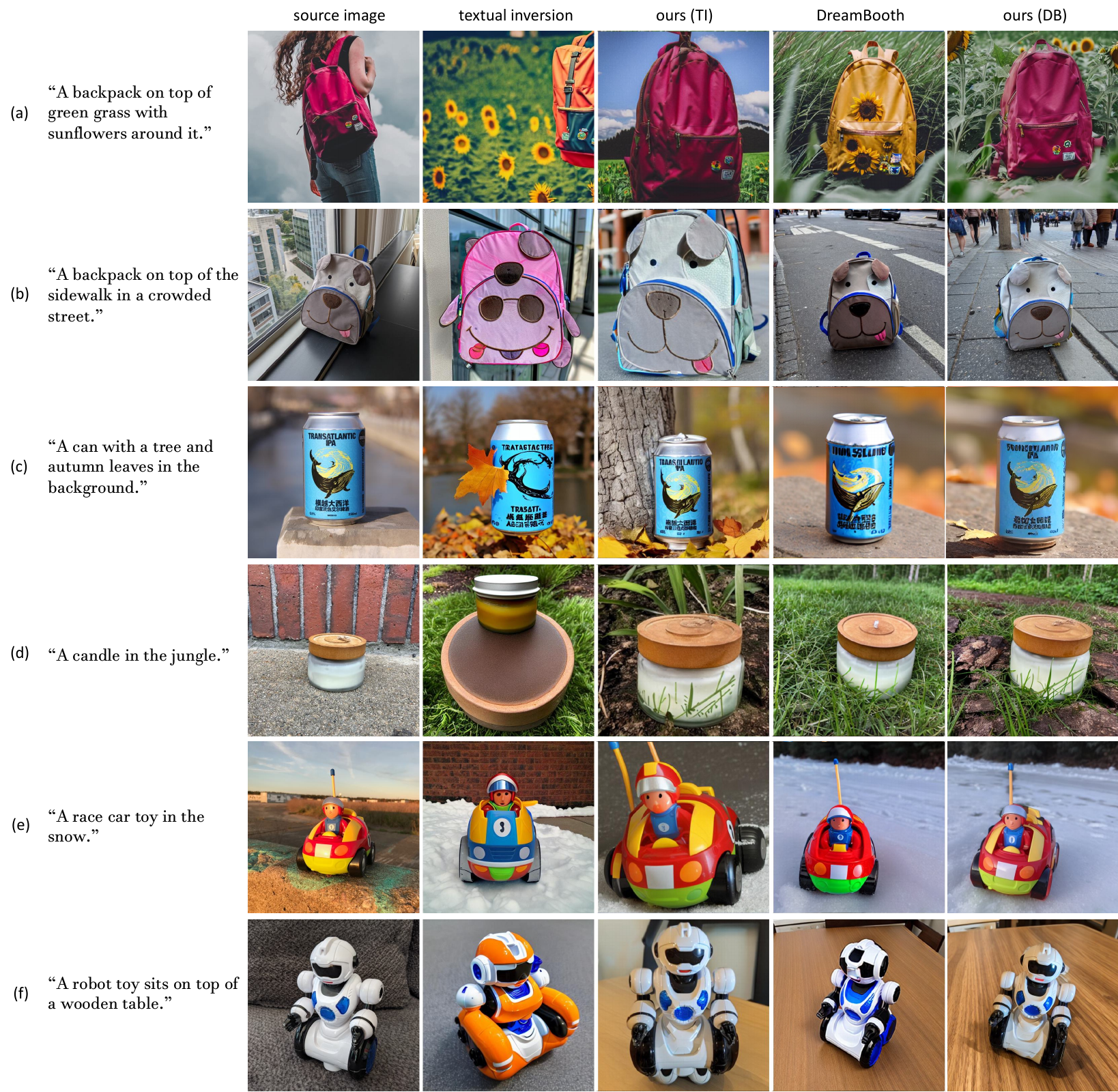}
\caption{\small{Qualitative comparison. We implement our fine-tuning method based on both Textual Inversion (TI) and DreamBooth (DB). A visible improvement is made by comparing the images in the third column with those in the second column and comparing the images in the fifth column and those in the forth column.}}
\label{pics:vis_comparison}
\end{figure*}

\begin{figure}[t]
\centering
\includegraphics[width=0.47\textwidth]{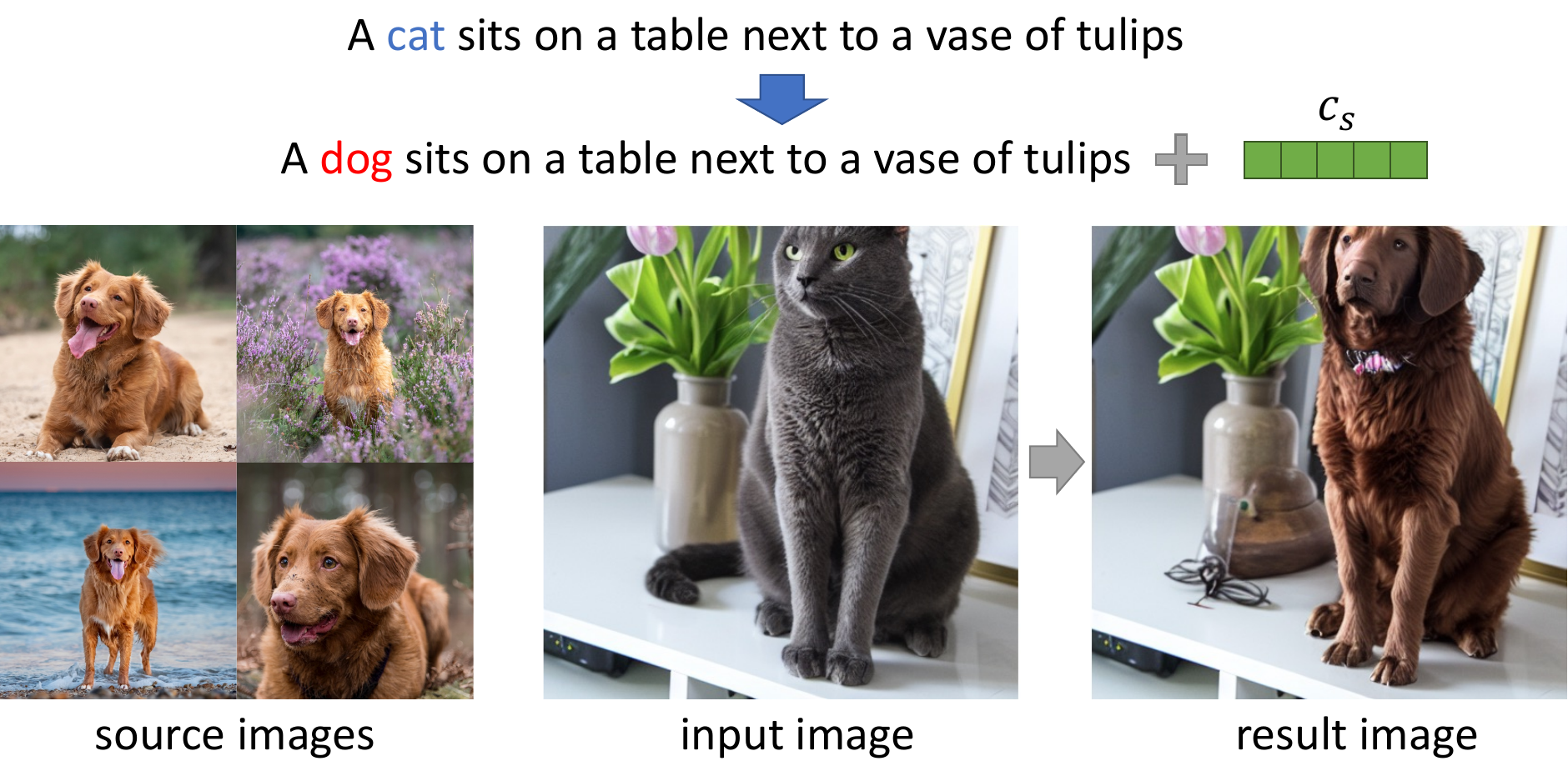}
\caption{\small{Results for personalized subject replacement.}}
\vspace{-3mm}
\label{pics:sub_sub}
\end{figure}

\textbf{Dataset}. We use the DreamBooth \cite{ruiz2023dreambooth} dataset for evaluation. It contains 30 subjects: 21 of them are rigid objects and 9 of them are live animals subject to large appearance variations. The dataset provides 25 prompt templates for generating images. Following DreamBooth, we fine-tune our framework for each subject and generate 4 images for each prompt template, totaling 3,000 images.

\textbf{Settings}. We adopt the pretrained Stable Diffusion \cite{rombach2022high} version 1.4 as the text-to-image framework. We use DDIM with 50 steps for the generation process. For HiFi Tuner based on Textual Inversion, we implement both the learning of subject textual embeddings described in Section \ref{subsec:denoise} and the reference-guided generation described in Section \ref{subsec:guided_infer}. We use 5 tokens for $c_s$ and adopts an ADAM \cite{kingma2014adam} optimizer with a learning rate $5e^{-3}$ to optimize it. We first optimize $c_s$ for 1000 steps and then recursively optimize $c_s^t$ for 10 steps per denoising step. We set $t_s=40$ and $t_e=10$ and use an ADAM \cite{kingma2014adam} optimizer with a learning rate $1e^{-2}$ to optimize $\phi_h$. We optimize $\phi_h$ for 10 steps per DDIM denoising step. For HiFi Tuner based on DreamBooth, we follow the original subject representation learning process and implement the reference-guided generation described in Section \ref{subsec:guided_infer}. We use the same optimization schedule to optimize $\phi_h$ as mentioned above. For the reference-guided generation, we only apply HiFi Tuner to the 21 rigid objects, because their appearances vary little and have strong need for the detail preservation. 

\textbf{Evaluation metrics}. Following DreamBooth \cite{ruiz2023dreambooth}, we use DINO score and CLIP-I score to measure the subject fidelity and use CLIP-T score the measure the prompt fidelity. CLIP-I score is the average pairwise cosine similarity between CLIP \cite{radford2021learning} embeddings of generated images and real images, while DINO score calculates the same cosine similarity but uses DINO \cite{caron2021emerging} embeddings instead of CLIP embeddings. As pointed out in the DreamBooth paper \cite{ruiz2023dreambooth}, DINO score is a better means than CLIP-I score in measuring the subject detail preservation. CLIP-T score is the average cosine similarity between CLIP \cite{radford2021learning} embeddings of the pairwise prompts and generated images.

\textbf{Qualitative comparison}. Fig.~\ref{pics:vis_comparison} shows the qualitative comparison between HiFi Tuner and other fine-tuning frameworks. HiFi Tuner possesses three advantages compared to other methods. First, HiFi Tuner is able to diminish the unwanted style change for the generated subjects. As shown in Fig.~\ref{pics:vis_comparison} (a) \& (b), DreamBooth blends sun flowers with the backpack, and both DreamBooth and Textual Inversion generate backpacks with incorrect colors; HiFi Tuner maintains the styles of the two backpacks. Second, HiFi Tuner can better preserve details of the subjects. In Fig.~\ref{pics:vis_comparison} (c), Textual Inversion cannot generate the whale on the can while DreamBooth generate the yellow part above the whale differently compared to the original image; In Fig.~\ref{pics:vis_comparison} (d), DreamBooth generates a candle with a white candle wick but the candle wick is brown in the original image. Our method outperforms Textual Inversion and DreamBooth in preserving these details. Third, HiFi Tuner can better preserve the structure of the subjects. In Fig.~\ref{pics:vis_comparison} (e) \& (f), the toy car and the toy robot both have complex structures to preserve, and Textual Inversion and DreamBooth generate subjects with apparent structural differences. HiFi Tuner makes improvements on the model's structural preservation capability. 

\textbf{Quantitative comparison}. We show the quantitative improvements HiFi Tuner makes in Table \ref{table:comparison}. HiFi Tuner improves Textual Inversion for 9.6 points in DINO score and 3.6 points in CLIP-T score, and improves DreamBooth for 1.2 points in DINO score and 1.2 points in CLIP-T score.

\begin{table}[t]
\small
    \centering
    \caption{\small{Quantitative comparison. }}
    \vspace{-8pt}
    \begin{tabular}{l|ccc}
        \toprule
         Method & DINO $\uparrow$ & CLIP-I $\uparrow$ & CLIP-T $\uparrow$  \\
         \midrule
         Real images & 0.774 & 0.885 & N/A \\
         Stable Diffusion & 0.393 & 0.706 & 0.337 \\
         \midrule
         Textual Inversion \cite{gal2023an} & 0.569 & 0.780 & 0.255 \\
         Ours (Textual Inversion) & 0.665 & 0.807 & 0.291 \\
         \midrule
         DreamBooth \cite{ruiz2023dreambooth} & 0.668 & 0.803 & 0.305 \\
         Ours (DreamBooth) & \textbf{0.680} & \textbf{0.809} & \textbf{0.317} \\
         \bottomrule
    \end{tabular}
    \label{table:comparison}
\end{table}

\begin{table}[t]
\small
    \centering
    \caption{\small{Ablation study.}}
    \vspace{-8pt}
    \begin{tabular}{l|ccc}
        \toprule
         Method & \footnotesize{DINO $\uparrow$} & \footnotesize{CLIP-I $\uparrow$} & \footnotesize{CLIP-T $\uparrow$}  \\
         \midrule
         Baseline (Textual Inversion) & 0.567 & 0.786 & 0.293 \\
         + mask & 0.606 & 0.788 & 0.292 \\
         + regularization & 0.612 & 0.789 & 0.294 \\
         + step-wise representations & 0.626 & 0.790 & 0.292 \\
         + reference guidance & 0.665 & 0.807 & 0.291 \\
         \midrule
         Baseline (DreamBooth) & 0.662 & 0.803 & 0.315 \\
         + reference guidance & 0.680 & 0.809 & 0.317 \\
         \bottomrule
    \end{tabular}
    \label{table:ablation}
\end{table}

\textbf{Ablation studies}. We present the quantitative improvements of adding our proposed techniques in Table \ref{table:ablation}. We observe that fine-tuning either DreamBooth or Textual Inversion with more steps leads to a worse prompt fidelity. Therefore, we fine-tune the networks with fewer steps than the original implementations, which results in higher CLIP-T scores but lower DINO scores for the baselines. Thereafter, we can use our techniques to improve the subject fidelity so that both DINO scores and CLIP-T scores can surpass the original implementations. For HiFi Tuner based on Textual Inversion, we fine-tune the textual embeddings with 1000 steps. The four proposed techniques make steady improvements over the baseline in DINO score while maintain CLIP-T score. The method utilizing all of our proposed techniques makes a remarkable 9.8-point improvement in DINO score over the baseline. For HiFi Tuner based on DreamBooth, we fine-tune all the diffusion model weights with 400 steps. By utilizing the reference-guided generation, HiFi Tuner achieves a 1.8-point improvement over the baseline in DINO score.

\textbf{Results for personalized subject replacement}. We show the qualitative results in Figure \ref{pics:sub_sub}. More results can be found in the Appendix ~\ref{app:person_sub_rep}.

\section{Conclusions}
\label{sec:conc}
In this work, we introduce a parameter-efficient fine-tuning method that can boost the sample fidelity and the prompt fidelity based on either Textual Inversion or DreamBooth. We propose to use a mask guidance, a novel parameter regularization technique and step-wise subject representations to improve the sample fidelity. We invents a reference-guided generation technique to mitigate the unwanted variations and artifacts for the generated subjects. We also exemplify that our method can be extended to substitute a subject in an image with personalized item by textual manipulations. 

{
    \small
    \bibliographystyle{ieeenat_fullname}
    \bibliography{main}
}

\clearpage
\appendix
\section{Failure analysis of Textual Inversion}
\label{app:TI_failure_analysis}

\begin{figure}[h]
\centering
\includegraphics[width=0.45\textwidth]{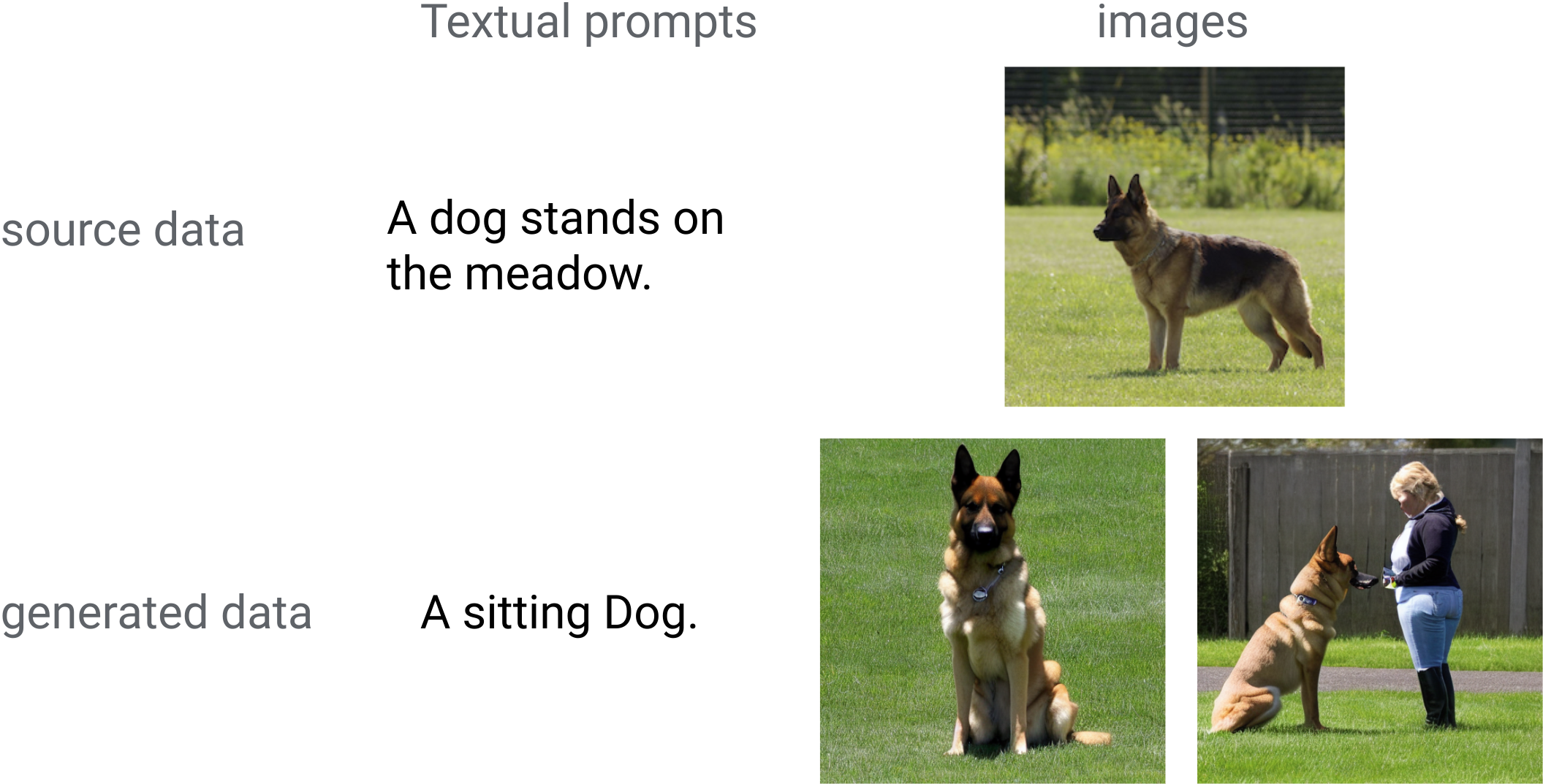}
\caption{\small{A failure analysis of Textual Inversion \cite{gal2023an} method.}}
\label{pics:TI_failure_analysis}
\end{figure}

We present a failure analysis of Textual Inversion as shown in Figure ~\ref{pics:TI_failure_analysis}. The source image is a dog sitting on a meadow. For a prompt “a sitting dog”, the generated images mostly contain a dog sitting on a meadow and the dog's appearance is not well preserved.

\section{Algorithm of optimizing $T_\theta$}
\label{app:opt_T_theta}
Please refer to Algorithm ~\ref{algr:opt_T_theta} for optimizing $T_\theta$.
\begin{algorithm}[h]
\SetAlgoLined
\small
 \caption{\small{Algorithm of optimizing $T_\theta$. $P(M) \in \mathbb{R}^{N\times3}$ returns the coordinates where $M==1$ and appends $1$'s after the coordinates.}}
 \label{algr:opt_T_theta}
\KwResult{$T_\theta^*$}
\textbf{Inputs}: $M_r$, $M_g$  \\
$P_r = P(M_r)$, $P_g = P(M_g)$ \\
\For{$l = [1,...,L]$}{
  $s = 0$ \\
  $P_t = T_\theta(P_r)$ \\
  \For{$p_t \in P_t$}{
    $m = MAX\_FLOAT$ \\
    \For{$p_g \in P_g$}{
      $x = \|p_t - p_g\|_2^2$ \\
      \If{$x < m$}{
        $m = x$ \\
      }
    }
    $s = s+m$
  }
  $\theta = \theta - \eta \nabla_{\theta} s$
}
$T_\theta^* = T_\theta$
\end{algorithm}

\section{Results for personalized subject replacement}
\label{app:person_sub_rep}
\begin{figure}[t]
\centering
\includegraphics[width=0.47\textwidth]{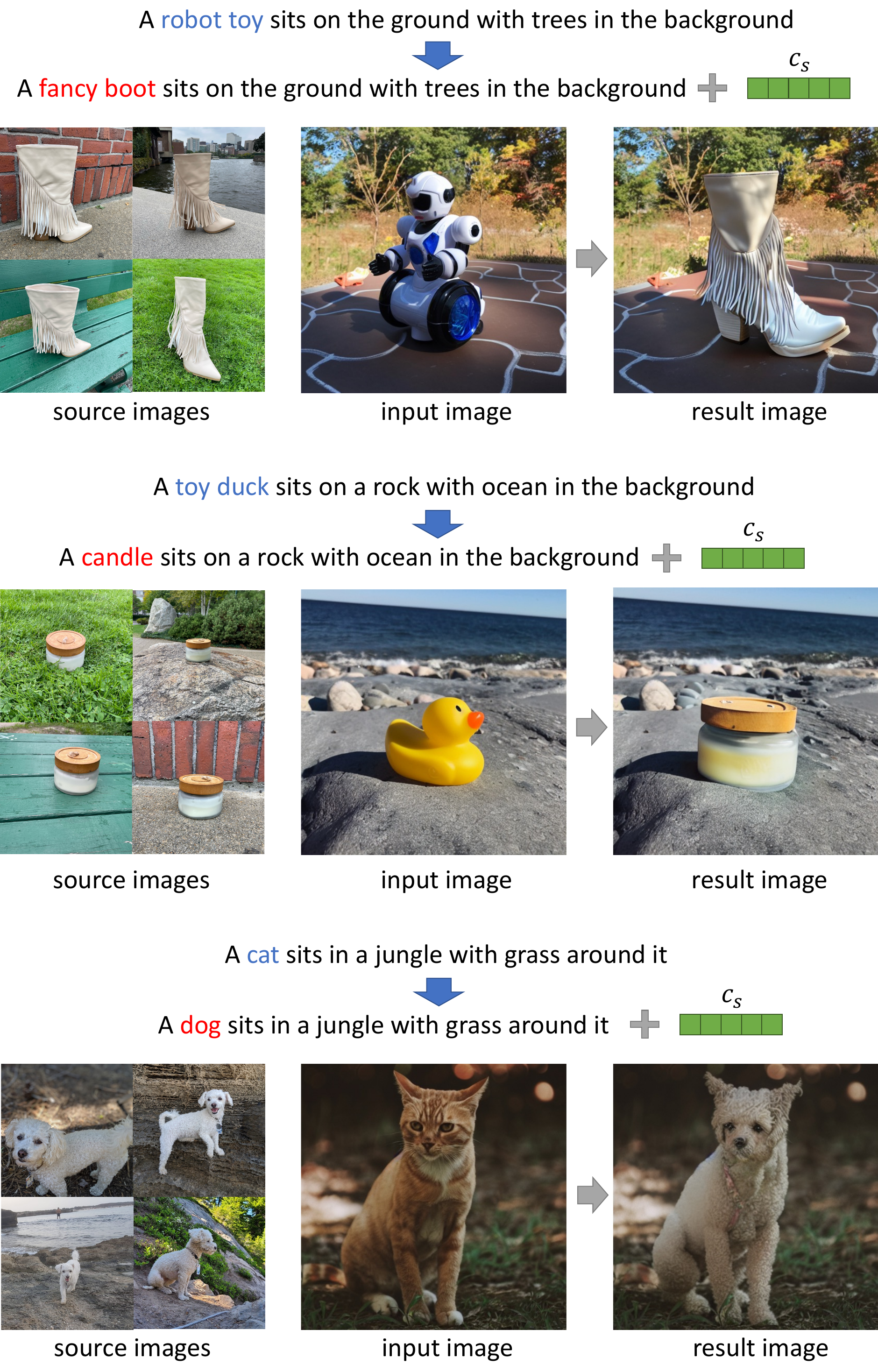}
\caption{\small{Results for personalized subject replacement.}}
\label{pics:sub_sub_sup}
\end{figure}

We show more results for the personalized subject replacement in Figure ~\ref{pics:sub_sub_sup}.

\end{document}